\documentclass[final]{cvpr}

\usepackage{times}
\usepackage{epsfig}
\usepackage{graphicx}
\usepackage{amsmath}
\usepackage{amssymb}
\usepackage{dirtytalk}
\usepackage{caption}
\usepackage{subcaption}

\usepackage[pagebackref=true,breaklinks=true,colorlinks,bookmarks=false]{hyperref}

\usepackage{pythonhighlight}

\usepackage{listings,lstautogobble}
\DeclareFixedFont{\ttb}{T1}{txtt}{bx}{n}{12} 
\DeclareFixedFont{\ttm}{T1}{txtt}{m}{n}{12}  

\usepackage{color}
\definecolor{deepblue}{rgb}{0,0,0.5}
\definecolor{deepred}{rgb}{0.6,0,0}
\definecolor{deepgreen}{rgb}{0,0.5,0}

\def\cvprPaperID{2} 
\def\confYear{CVPR 2021}

\let\oldquote\quote
\let\endoldquote\endquote
\renewenvironment{quote}[2][]
  {\if\relax\detokenize{#1}\relax
     \def\quoteauthor{#2}%
   \else
     \def\quoteauthor{#2~---~#1}%
   \fi
   \oldquote}
  {\par\nobreak\smallskip\hfill(\quoteauthor)%
   \endoldquote\addvspace{\bigskipamount}}

\newcommand\blfootnote[1]{%
  \begingroup
  \renewcommand\thefootnote{}\footnote{#1}%
  \addtocounter{footnote}{-1}%
  \endgroup
}

\begin{document}

\title{Continuum: Simple Management of Complex Continual Learning Scenarios}

\author{Arthur Douillard\footnotemark\\
Sorbonne Université, Paris\\
Heuritech\\
{\tt\small arthur.douillard@heuritech.com}
\and
Timothée Lesort\footnotemark[\value{footnote}]\\
Université de Montréal, Mila - Quebec AI Institute\\
{\tt\small timothee.lesort@mila.quebec,}
}

\maketitle

\blfootnote{$^*$ Equal Contribution}

\begin{abstract}
Continual learning is a machine learning sub-field specialized in settings with non-iid data.
Hence, the training data distribution is not static and drifts through time. Those drifts might cause interferences in the trained model and knowledge learned on previous states of the data distribution might be forgotten. Continual learning's challenge is to create algorithms able to learn an ever-growing amount of knowledge while dealing with data distribution drifts.

\noindent One implementation difficulty in these field is to create data loaders that simulate non-iid scenarios. Indeed, data loaders are a key component for continual algorithms. They should be carefully designed and reproducible. Small errors in data loaders have a critical impact on algorithm results, e.g. with bad preprocessing, wrong order of data or bad test set.
\textbf{Continuum} is a simple and efficient framework with numerous data loaders that avoid researcher to spend time on designing data loader and eliminate time-consuming errors. Using our proposed framework, it is possible to directly focus on the model design by using the multiple scenarios and evaluation metrics implemented.
Furthermore the framework is easily extendable to add novel settings for specific needs.

\end{abstract}

\section{Introduction}

\begin{figure}
     \begin{subfigure}[b]{0.5\textwidth}
         \includegraphics[width=\textwidth]{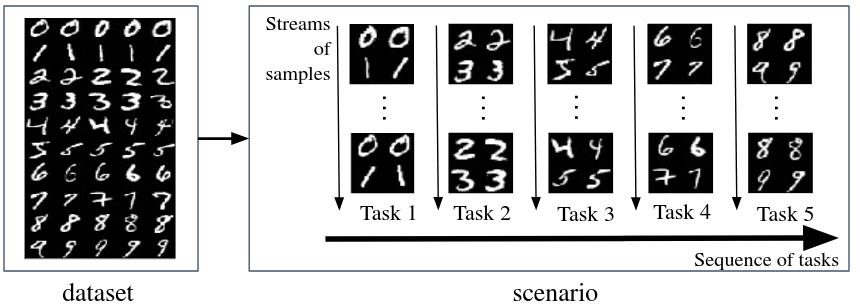}
         \caption{Class-Incremental MNIST with 5 tasks of 2 new classes each.}
         \label{fig:mnist}
     \end{subfigure}
     \hfill
     \begin{subfigure}[b]{0.5\textwidth}
        \begin{python}
        # create a scenario from a set of
        scenario = ClassIncremental(dataset, increment=2)
        
        # load tasks one by one
        for task in scenario:
            loader = DataLoader(task)
            # create the task stream of samples
            for samples, labels, task_ids in loader:
                # Learn here
        \end{python}
        
         \caption{Pseudo-Code for class-incremental scenario with 2 classes increment.}
     \end{subfigure}
\caption{Class-incremental scheme with the associated pseudo-code to present Continuum architecture to create continual scenarios. More complex scenarios are also available.}
\label{fig:code_presentation}
\end{figure}




In deep learning, the implementation of algorithms and their debugging takes an important part in the research time. Therefore, numerous types of libraries have been developed to help researchers to quickly implements their algorithms. Those libraries might be very low level for optimized use of hardware such as Cudnn \cite{Chetlur2014Cudnn} or Cuda \cite{cuda} or it might also help to optimized mathematical operation such as NumPy \cite{harris2020array}. There are also higher-level libraries that gathered together a set of tools for research. In machine and deep learning, multiple libraries such as scikit-learn \cite{scikit-learn}, Pytorch \cite{NEURIPS2019_9015} or TensorFlow \cite{tensorflow2015-whitepaper} ease the conception of learning algorithms.

While it is relatively easy to implement data loaders in the supervised learning with high-level aforementioned frameworks, it is considerably harder for continual learning. The non-iid property imposes to careful build data loaders whose behaviors change through time. In this paper, we propose a library, \textbf{Continuum}, that propose easy-to-use loaders for such non-iid scenarios. Following the Unix philosophy, our library aims to do few things, data loading and metrics, but to do it well and with minimal assumptions in order to let researchers implement their own  algorithms. Our framework is designed to be simple enough in order to being adaptable to any specific needs and other libraries: figure \ref{fig:code_presentation}, introduces the basic architecture of continuum in a few lines of code.

We will first motivate the need of this library by presenting the main continual learning scenarios and metrics in the literature. In a second time, we will present the different application fields of continual learning. To finish, we will present how continuum is structured to fit the various scenarios and applications and we showcase code examples.

\section{Continual Learning Scenarios}
\label{sec:continual_scenarios}

Continual learning scenarios are various, they might be characterized by the evolution of their data distribution and by their supervision specificities. 
In this section, we present the main characterization of continual learning scenarios. Moreover, we describe the evaluation metrics usually used in continual learning.

\subsection{Data Variations Specificities}

Catastrophic forgetting is caused by variation in the data distribution. Those variations might be used to classify the continual scenarios. We list here continual learning types of scenarios associated with specific types of data variation.

\subsubsection{Incremental learning (New task $\rightarrow$ New concepts)}

\begin{figure}
\centering
  \includegraphics[width=0.7\linewidth]{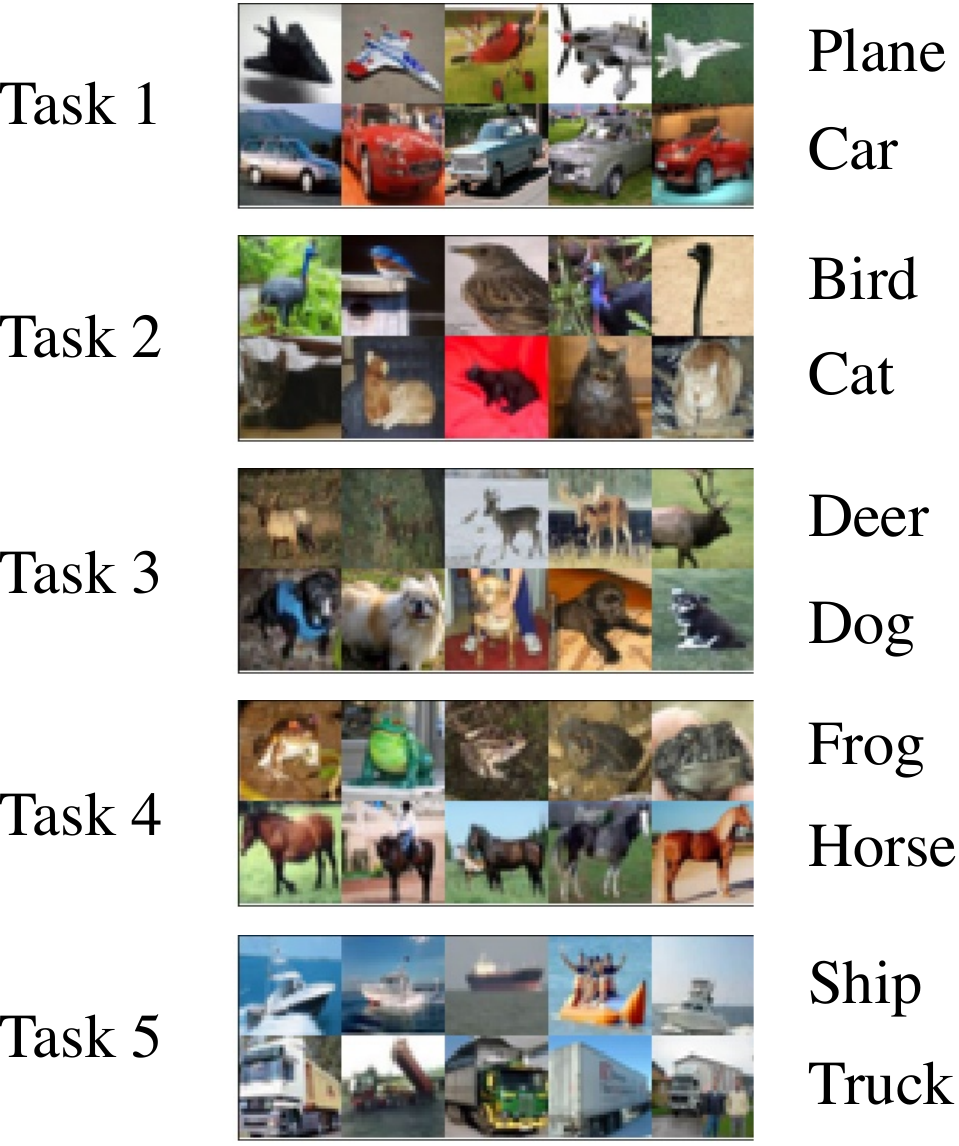}
\caption{CIFAR10 learned in a class-incremental fashion in 5 tasks of 2 new classes each.}
\label{fig:cifar10}
\end{figure}

Incremental learning is about learning new concept sequentially and being able to retain all those concepts in a single model. 
We refer to a concept, as a labeled object. Two different concepts are then two objects labeled differently.
In classification, this paradigm is incarnate by scenario when new tasks bring new classes. At the end of a sequence of 5 tasks containing 2 classes each, the model should be able in the end to classify an object from 10 classes. Figure \ref{fig:cifar10} illustrates such scenario. In reinforcement learning, it can be seen as a multi-tasks settings were tasks are learned one by one, e.g. learning to play Space Invader then Quest, then Pacman.

These scenarios evaluate the ability of a model to learn new concepts without forgetting previous ones and the ability to learn to differentiate concepts that are not available at the same time, e.g. learning the difference between classes from a task $t$ and a task $t+2$.

\subsubsection{Lifelong learning (New task $\rightarrow$ New instances, same concepts)}

Lifelong learning is about keeping improving at a certain task with new data. For example, in classification, this paradigm is incarnated by a sequence of tasks with the same classes but different data points. It is particularly interesting when the new data points come from a modified distribution, e.g. cow on grass, cow on sand beaches, etc. It is also called domain incremental or instance incremental.

It evaluates the capacity of a model to improve its known concepts with new data. We can interpret it at using new data to improve the latent representations of objects and improve generalization.

\subsubsection{Others}

First, Incremental learning and lifelong learning are not incompatible and can be mixed to evaluate both the capacity of a model to improve on known concepts and add new concepts. This scenario is also known as \say{New Instances and Classes} scenarios (NIC) \cite{lomonaco2017core50, lesort2020continual}.

Secondly, a different variation may happen in the data distribution. Indeed, the label distribution might drift. In other words, a given sample may have various label through time. This type of drift is called real concept drift \cite{gama2014survey,gepperth2016incremental}.
In this scenario, the model has to choose between learning the various labels for one input or to forget past labels to only remember the current label and be up to date.

\medskip

In a sens, incremental and lifelong learning are fundamentally different. Indeed, incremental learning is about learning to distinguish concepts from different data distributions and lifelong learning is about learning the similarity between different data distributions.
Algorithms are not necessarily able to do both and therefore distinguishing them enable more precise evaluation of an algorithm's capabilities.

\subsection{Supervision Specificities}

The learning ability of algorithms is largely dependant on supervision.
The supervision specificities characterize the information we have about the data, e.g. their labels. Moreover, it may inform on the data distribution changes. In this section, we present how those distribution changes might be labeled and discuss some assumptions related to continual learning scenarios.

\subsubsection{Task labelization}

In continual learning, we often split the data stream into subparts we call tasks. The task are usually, but not necessary, subsets where the data distribution can be assumed i.i.d.. In this case, a task change when the data distribution change. It is important to note that the task are defined empirically for learning convenience.
A task label is associated to each task and informs the learning algorithm about the task evolution \cite{Lesort2019Continual}. 
We distinguish two kind of tasks labels:

\begin{itemize}
    \item \textbf{Train task labels}: This label is given while learning and helps the algorithms to detect drift in data distribution and avoid interferences. In incremental learning, this task label is implicit.
    \item \textbf{Test task labels}: This label is given for inference, it indicates to the model from which task a data point is coming from.
\end{itemize}

The task labelization at test time is for example crucial for multi-head architecture while it is not necessary for single head architecture. Test task labels may be seen as a supervision signal for inference.
The only case where a test task label is absolutely necessary is when the algorithm needs context to know what is the current task to solve. For example, a robot that learns to walk and to jump, at test time, we should use the test task label to tell the robot what is the politic to run.

\medskip

The task label enables the algorithms to identify changes in the data distribution and associates input to a task. Like any type of supervision, it helps the algorithm to learn a valid solution to a problem but it is costly and might be unable in certain settings. The availability of tasks label can also be sparse and signal only a few data distribution changes.

\subsubsection{Questioning our assumptions}


Despite the variety of continual settings previously described, they can not be fully described by their labelization specificities and data distribution drifts. 
The set of assumptions needed for a model to learn are crucial for the success of an approach and they might vary from one setting to another, as well as from a learning approach to another. 
Those assumptions are made by the algorithms designer about the learning environment and usually help algorithms to solve their tasks. Therefore, they can be seen as (weak) supervision information.

For example, many assumptions characterize the data stream and the information algorithms get about it. In continual learning, numerous approaches assume a fixed labelization and uncorrelated tasks.

However, recent works challenge some of those common assumptions such as fixed label \cite{cermelli2020modelingthebackground,abdelsalam2021iirc}, uncorrelated data stream \cite{caccia2020online,smith2021semisupcontinual}, labeled new data \cite{hanrebuffi2020autodiscovering}, or even no meta-data about tasks and unavailability of information about the future \cite{douillard2020ghost}.


Continual learning is, by definition, linked to the data distribution evolution. Our assumptions or knowledge about it, can be various and used in multiple manners. 
Hence, a clear description of setting assumptions used in an approach is crucial to understand algorithms supervision needs as well as for reproducibility and evaluation.

\subsection{Metrics}



In order compare clearly different continual learning approaches, we must evaluate them with the right metrics. While the accuracy metric is predominant, multiple alternatives exist: some to quantify raw performance, others to explain the behavior of the models. Most of these metrics are peculiar to the domain of Continual Learning because they measure a succession of models instead of a unique model.


\subsubsection{Performance Metrics}

Because of the iterative nature of Continual Learning scenarios, multiple metrics exist to evaluate model performance. Intuitively, the accuracy of the model after the last task ($A^t$) is important. However, it's not sufficient, e.g. two different models could have the same accuracy at the task $t$ but one of them could have better accuracies at task 1, 2, ..., $t-1$. Rebuffi et al.~\cite{rebuffi2017icarl} proposed then the \texttt{average incremental accuracy} which is the average of the accuracies after each task: $\frac{1}{T} \sum_{t=1}^T A^t$.

The model can also be evaluated by its performance while learning, for example is the online cumulative performance \cite{caccia2020online}.

\subsubsection{Auxiliary Metrics}

Several additional metrics were created for continual learning. They can be sundered into two categories: (1) \textit{behavior} and (2) \textit{computational}. 

\noindent\textbf{Behavior metrics} provide additional insights on the model learning. e.g. the \texttt{backward transfer} \cite{lopezpaz2017gem} measures the influence that learning a task has on previous tasks. This metric was latter split into \texttt{remembering} and \texttt{positive backward transfer} by \cite{diaz2018metriccontinual} to record the forgetting and improvement brought by new tasks.

\noindent\textbf{Computational metrics} record the impact of the training on hardware. Particularly important on embedded devices with limited power, it measures the memory consumption ---from both the models and stored samples, the number of flops, etc. For example Progressive Networks \cite{Rusu16progressive} showcase strong resilience to forgetting but its memory consumption is also prohibitively high due to the cost of storing all previous models.

\medskip

We have seen that various types of scenarios, supervision and evaluation might be used in continual learning. Continuum was build to gather together the various cases in a single library and ease their use.

\section{Application Fields}
\label{sec:applications}

In the previous section, we have seen the several types of continual learning scenarios, their supervision specifities and evaluation metrics. In this section, we present the various applications fields to continual learning. The goal is to show that the library we propose could be used in many types of learning problems.

\subsection{Image Classification}


Currently, the majority of Continual Learning models are applied in the context of image classification. 
Using classification settings is convenient because most of the data are annotated. It makes possible to study continual learning problems, such as catastrophic forgetting without interfering with other machine learning problems (e.g. sparse labelization, reinforcement learning...). As well-annotated data eases the training on a task, the definition of tasks and the continual evaluation can be more clearly investigated. Although note that, while the most common approach in continual image classification is based on full supervision, semi-supervised approaches also exist \cite{smith2021semisupcontinual}.

A common setting in continual image classification is a dataset whose $N$ classes are split across $T$ tasks. e.g. during the first task, the continual model learns to classify $\frac{N}{T}$ classes. The second task brings $\frac{N}{T}$ classes, etc. This scenario belonds to the incremental learning group.


This kind of scenario might be easier to solve than most of reinforcement learning scenarios but they stay plausible for real life application and makes clearer evaluations of algorithms performances. Therefore, they are highly valuable for continual learning research community.


\subsection{Reinforcement learning}

In reinforcement learning, the data distribution is rarely i.i.d. because the training data distribution depends on the policy. Continual learning is therefore naturally adapted to this learning paradigm \cite{khetarpal2020continual}.

Two kind of scenarios can arise in Reinforcement Learning. Either our model has to learn sequentially several environment (i.e. space invader, then quest, then pacman, etc.) \cite{kirkpatrick2017ewc,jung2020groupsparsereg} or tasks  \cite{Kalifou19,Traore19DisCoRL},  (incremental learning) either the current environment change while learning a single tasks (lifelong learning).
From the perspective of the algorithm, the data distribution always change as the policy change. Therefore, it is not easy for it to tell apart changes due to the policy update or due to a new task or a new environment. 
Nevertheless, Continual RL is more associated to case were the training data is not i.i.d \textbf{and} where their is several type of agent, environment or tasks.

Solving reinforcement learning scenarios helps to push research in situation where there are few feedbacks from the environment.


\subsection{Continual Semantic Segmentation}

Continual Semantic Segmentation \cite{michieli2019ilt} is the combination of continual learning and semantic segmentation \cite{everingham2015pascalvoc,zhou2017adedataset,lin2014mscocodataset,cordts2016cityscapes}. This setting has been refined into two scenarios: Disjoint and Overlapped \cite{cermelli2020modelingthebackground,douillard2020plop}. In both scenarios, classes are added incrementally and no storing of previous images is allowed. In the former, an image of task $t$ may contain either old or current classes. In the latter, it can also contain future yet-to-be-seen classes. In both scenarios, the ground-truth segmentation masks are partially labeled: current classes are labeled but all others classes are assigned a background label. The goal of this setting is therefore to learn continuously new classes, while not forgetting old classes that may be present in the current images but unlabeled.

This setting has real-life applications such as autonomous driving where fully labeling the segmentation mask of a large image is time consuming and cost expensive. 

\subsection{Natural Language Processing}

Continual learning is not restricted to computer vision. Asghar et al.~\cite{asghar2020nlpincrementalmemorybanks} proposed to solve the task of Natural Language Inference (NLI) when faced to an incremental stream of task datasets. Each dataset is a text corpus belong to particular domain \cite{bowman2018multinli}; e.g. fiction, letters, face-to-face, etc. This application belongs to the lifelong learning group where the concepts are kept fixed but the input distribution evolves through time. More recently Lovon et al.~\cite{lovon2021continualri} explored the domain of information retrieval when faced to a stream of multiple text sources.  While still minor compared to computer vision, we believe that continual learning transcends data types and many existing approaches can be applied to natural language processing with few modifications.

\medskip

We did not list all the application fields where continual learning could be used but there are many others such as regression or data generation. Continuum is designed to be as agnostic as possible to the data type in order to be adapted to any of those application fields easily. In fact, Continuum already proposes Asghar et al.~\cite{asghar2020nlpincrementalmemorybanks} setting.

\section{Continuum}

In this section, we introduce how the continuum library could be used to create scenarios compatible with the use cases and settings presented in the previous sections. In the existing continual learning, many research papers share their research code and their environment such as \cite{Lopez-Paz17, lomonaco2017core50, riemer2018learning, lesort2018generative, deLange2019continual, aljundi2019online,douillard2020podnet}. However, their code base is not necessary designed to be adapted easily to other settings and contains only the scenarios of their experiments. 
Continuum is therefore developed to gathered most of the existing scenarios into one single codebase that could be easily used, extended and adapted.
Hence, \textit{Continuum} is an open-source framework to design and use continual learning scenarios 
\footnote{Code available at \url{https://github.com/Continvvm/continuum}.}. 
It was created to provide pre-made scenarios to continual learning researchers. The goal is to gain time and avoid bugs in a continual learning project. It is also made to improve reproducibility of results by providing normalized scenario settings. We wish to stress that the philosophy of Continuum is to be easily integrated to any continual learning projects irrespective of the models used. In this section, we will present the architecture of continuum to create scenarios. Moreover, we introduce the metrics that are provided by the framework to evaluate algorithms.

\subsection{Organization}

\begin{figure}
\centering
  \includegraphics[width=\linewidth]{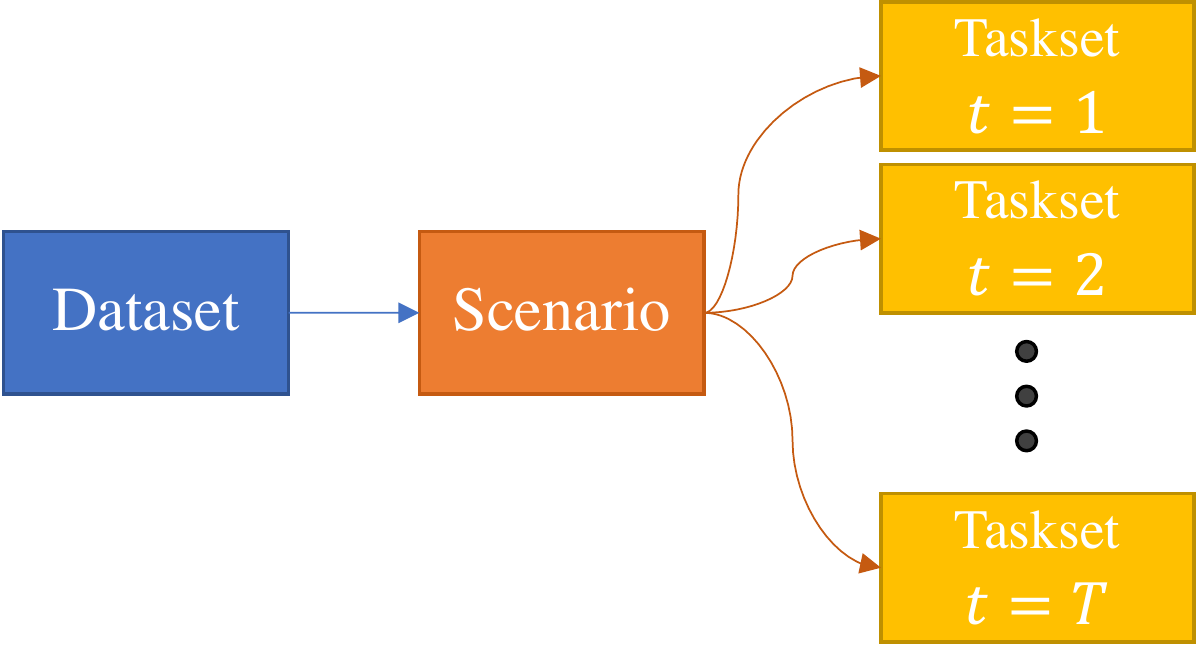}
\caption{Organization of the three main blocks of Continuum: a \texttt{Scenario} takes a \texttt{Dataset} as input and produces a sequence of \texttt{TaskSets}.}
\label{fig:continuum_codebase}
\end{figure}

To create continual learning scenarios, Continuum decomposes the data management into three levels of data structures: Datasets, Tasksets, Scenarios.

\begin{itemize}
    \setlength\itemsep{0.05em}
    \item \textbf{Datasets:} Datasets are the raw data that will be used to create tasks and scenarios. Most proposed datasets can automatically download\footnote{At the exception of datasets where data access necessitates login.} and format the data directly for the continual tasks.
    \item \textbf{Tasksets:} The taskset contains the data specific to a task. The data are selected from the original dataset and eventually transformed.
    \item \textbf{Scenarios:} A scenario is a sequence of tasks. It composes the curriculum of learning experience fed to the algorithms.
\end{itemize}

This decomposition makes possible to change independently the data types (Datasets), the data distribution drifts and task labels (Scenarios) and the data stream in a single task (Taskset). This structure makes possible to create any use case presented in the sections \ref{sec:continual_scenarios} and \ref{sec:applications}. For example, continuum can be used to create one continuous stream of data for online learning or several sequences with clear distinction for incremental learning.

The user only has to specify the dataset and the scenario, and the different tasks sets will be created automatically as illustrated in Figure \ref{fig:continuum_codebase}.

\subsection{Scenarios}

\begin{figure}
\centering
  \includegraphics[width=\linewidth]{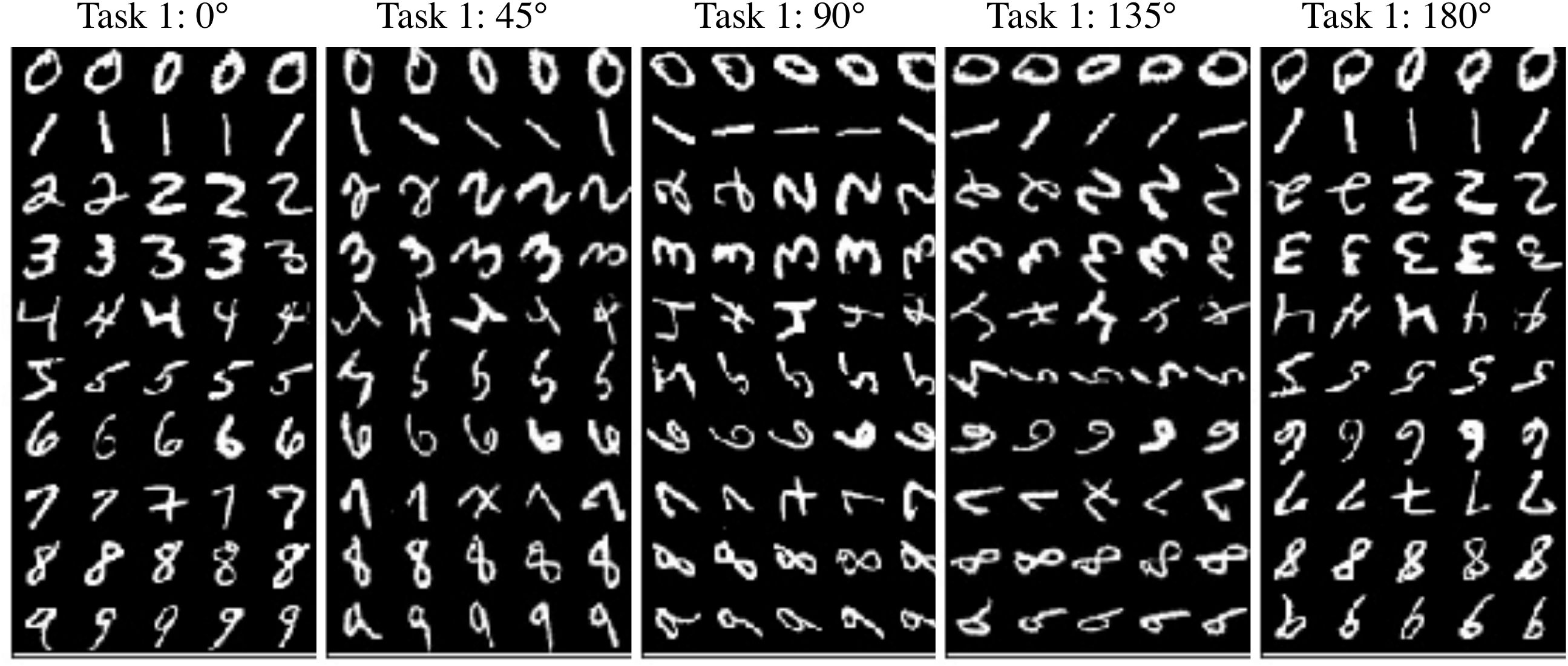}
\caption{An example of Transformation-based dataset: Rotation-MNIST where each task bring the same images but with 45° rotation relative to the precedent task.}
\label{fig:mnist_rot}
\end{figure}

The scenario defines how the dataset is decomposed into several tasks. Continuum provides several pre-made scenarios that often occur in the literature. As defined in \autoref{sec:continual_scenarios}, \texttt{Incremental Learning}, \texttt{Lifelong Learning}, and a combination of both are provided. We also provide a transformation scenario where each new task is created by applying a different transformation to the dataset: two common examples are Rotation-MNIST (featured in Figure \ref{fig:mnist_rot}) and Permuted-MNIST.

\subsection{Supported Datasets}

In \textbf{Continuum}, there are three kinds of dataset:

\begin{itemize}
\setlength\itemsep{0.05em}
 \item Standard datasets (i.e. MNIST, ImageNet, etc.)
 \item Transformation-based datasets where a particular transformation is applied at different step (i.e. Permuted-MNIST, Rotation-MNIST, etc.)
 \item Datasets with additional meta-data
\end{itemize}

The latter comprises datasets such as CORe50 where each sample, in addition of having a label, has a specified id that is used to build scenarios that bring both new classes and new samples (nicknamed NIC in \cite{lomonaco2017core50}). 


Continuum supports all the basic datasets from \texttt{pytorch.datasets} (MNIST \cite{lecun2010mnist}, CIFAR10, CIFAR100 \cite{krizhevskycifar100}) as well as larger datasets such as ImageNet or CORe50 \cite{lomonaco2017core50}. We provide also tools to create new datasets. For example, the fellowship class make possible to concatenate several datasets into one for specific scenarios. You can find a complete list of supported datasets on \url{https://continuum.readthedocs.io}.

\subsection{Code Examples}

Setting up a Continual Learning dataset is simple with Continuum. Figure \ref{fig:code_loader} showcases how to create a Split-MNIST scenario where each task is composed of 2 classes. Since MNIST is composed of 10 classes, it will create a scenario of 5 tasks.

\begin{figure}
\begin{python}
from torch.utils.data import DataLoader

from continuum import ClassIncremental
from continuum.datasets import MNIST

dataset=MNIST(
    "my/data/path",
    download=True,
    train=True
)

scenario = ClassIncremental(
    dataset,
    increment=2
)

print(f"Nb classes: {scenario.nb_classes}.")
print(f"Nb tasks: {scenario.nb_tasks}.")

for task_id, taskset in enumerate(scenario):
    loader = DataLoader(taskset)
    for x, y, t in loader:
        # Train your model here
\end{python}
\caption{Code example of a class-incremental scenario with the MNIST dataset \cite{lecun2010mnist}.}
\label{fig:code_loader}
\end{figure}

All commonly used metrics in Continual Learning are implemented and provided through a class \texttt{Logger}. After each step (i.e. when a task is finished) or after each batch, simply feed the logger with the model's predictions, the actual labels, and their associated task ids as shown in Figure \ref{fig:code_metrics}.

\begin{figure}
\begin{python}
from continuum import Logger

logger=Logger()
logger.add_step(
    predictions,
    labels,
    task_ids,
    model=model
)

print(logger.accuracy)
print(logger.remembering)
print(logger.model_size_efficiency)
\end{python}
\caption{Code example of the recording of model predictions and the resulting metrics scores.}
\label{fig:code_metrics}
\end{figure}

\medskip

In this section, we presented an overview of how continuum has been organized and the various settings and metrics it proposes to avoid the data processing work for the researcher.
Continuum has also been designed to be easy to use for any learning algorithms. The code organization is made to be easily adaptable and each object can be inherited to create more complex or specific scenarios without rewriting everything.

\section{Discussion}

At the time of writing this article, the choice has been made to not incorporate reinforcement learning (RL) environment into the continuum environment. This choice has been made because we believe that there are a lot of RL platforms allowing to train agent continually on or several tasks. Those existing platforms such as stable-baselines \cite{stable-baselines,stable-baselines3} makes easy to create sequences of settings or environment to train a reinforcement learning agent, that are sufficient, from our point of view, for continual RL.

At the moment, most datasets proposed in \textit{Continuum} are image classification based. We plan to add other type of datasets such as segmentation and detection \cite{everingham2015pascalvoc,lin2014mscocodataset,cordts2016cityscapes}. We also plan to increase our text-based datasets portofolio and to provide audio-based datasets.

\section{Conclusion}

Continuum is an open-source project which aims at simplifying data management for continual learning algorithms. 
It aims at covering various type of scenarios in continual learning research field and it is developed such as being easily adaptable to specific needs. 
%
Continuum is made to save development time, reduce code size in continual project, and avoid classical data loader bugs.

Moreover, we believe that the use of a common plateform for continual learning experiments will benefit to the community by facilitating reproducibility. 


{\small
\bibliographystyle{ieee_fullname}
\bibliography{egbib,continual,others}
}

\end{document}